%% file: main.tex
\documentclass{article}
\usepackage{iclr2027_conference,times}

\usepackage{amsmath}
\usepackage{amssymb}
\usepackage{array}
\usepackage{xcolor}
\definecolor{alphapink}{HTML}{B36680}
\definecolor{betabluebase}{HTML}{4D78A8}
\colorlet{betablue}{betabluebase!80!gray}

\input{math_commands.tex}

\usepackage[hypertexnames=false,hidelinks]{hyperref}
\usepackage{url}
\definecolor{wmlinkcolor}{RGB}{40,80,150}
\usepackage{graphicx}
\usepackage{float}
\usepackage{algorithm}
\usepackage{algpseudocode}
\algrenewcommand\algorithmicrequire{\textbf{Input:}}
\usepackage{tikz}
\usetikzlibrary{positioning, arrows.meta, calc, shapes.geometric, fit, backgrounds, decorations.pathreplacing}
\usepackage{microtype}
\usepackage{enumitem}
\usepackage{capt-of}
\usepackage{wrapfig}

\input{figures/architecture_4panel}

\input{figures/pool_pipeline}

\input{figures/iteration_modes}

\title{WhiteMatter: All-to-All Cross-Layer Connections via KV Source Mixing}

\author{Wenbo Zhang~~~~Xiang Ren \\
University of Southern California \\
\texttt{\{wenboz,xiangren\}@usc.edu}
}

\iclrpreprintcopy
\begin{document}

\maketitle

\begin{abstract}
  When generating text, a Transformer produces representations of past tokens at every layer, but each attention layer can normally use only representations from the same depth. This restriction prevents the model from fully reusing information it has already computed. We introduce \textbf{WhiteMatter}, which allows every layer to draw on past-token representations from any depth. A learned mixer selects the most useful depths for the current context and combines their representations into shared key--value (KV) cache channels. Sharing these channels across layers can reduce the cache size. Given the same number of training tokens, WhiteMatter with a full-size cache performs comparably to a standard Transformer with $50\%$ more layers. With half the KV cache, WhiteMatter outperforms matched standard Transformers at two model scales, up to $1.3$B parameters. Cross-layer connections, however, introduce dependencies that slow training and prompt processing. We address this problem with cyclic iteration, which updates interleaved groups of tokens in turn while processing the tokens within each group in parallel. On a reference model trained with exact autoregressive execution, cyclic iteration converges $12.5\times$ faster than standard Jacobi iteration.

  Code: \url{https://github.com/Cy-47/White-Matter}
\end{abstract}

\section{Introduction}

At each decoding step, a Transformer~\citep{vaswani2017attention} processes the current token through every layer before moving to the next token. Yet
each layer attends to earlier tokens using only keys and values (KV) produced
at the same depth. This restriction makes training and prompt processing easy
to parallelize, but prevents layers from reusing representations already
computed at other depths. As demands grow, inference can cost
more than training~\citep{openai2020efficiency}. In particular, decoding can often dominate inference time for agents and when generating
long reasoning traces~\citep{saxena2025utility,yuan2026agentic}. It is therefore important to relax this restriction while retaining the practical benefits of the
standard Transformer.

\begin{figure*}[b]
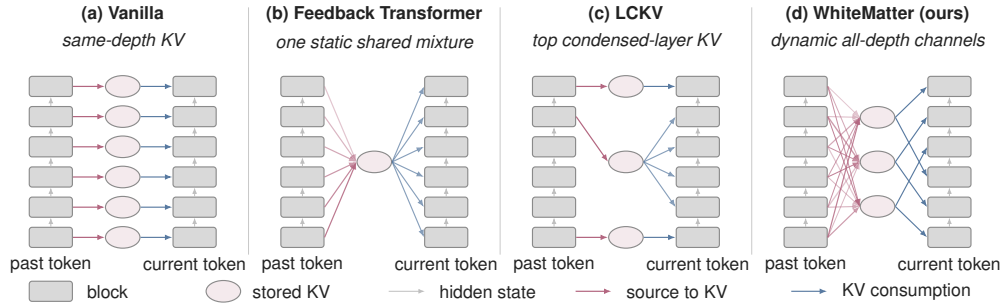

  \vspace{-0.0cm}
  \centering
  \resizebox{0.96\linewidth}{!}{\architecturefourpanelfigure}
  \vspace{-0.2cm}
  \caption{\textbf{KV production and consumption across layers.}
    \textbf{(a)} Vanilla: each layer reads KV produced at the same depth.
    \textbf{(b)} Feedback Transformer: all layers read KV produced from the same weighted sum of source depths.
    \textbf{(c)} LCKV: warmup layers retain
    same-depth KV, while condensed layers read KV produced from the top layer in the block.
    \textbf{(d)} WhiteMatter: a router forms $k$ channels from
    all source depths, and each layer reads one of the channels.}
  \label{fig:architecture}
  \vspace{-10px}
\end{figure*}

Deep-to-shallow feedback offers one way to reuse more of this computation. Feedback Transformer~\citep{fan2020feedback} combines all layer states for each past token into one weighted sum, then projects that summary into KV shared by every attending layer. LCKV~\citep{wu-tu-2024-layer} instead builds
shared KV from the deepest state, which has accumulated information from earlier layers through residual connections. LCKV performs better when some
``warmup'' layers retain their own same-depth KV, whereas applying feedback KV to every layer substantially worsens perplexity~\citep[Figure~8]{wu-tu-2024-layer}. This raises a central question: why does broader feedback become less effective when it is used everywhere?

One likely reason is the shared bottleneck in both methods: they compress
the available source states into a single layer-width representation used by
every target layer. Such a summary must
preserve every feature that any layer may need. In practice, intermediate
states can remain useful even after deeper layers have transformed them, and
our ablations show that simply enlarging the KV cache is not enough. We therefore hypothesize that different
target layers need access to different source depths. LCKV's warmup layers
partly restore intermediate information, but only through the same-depth
connections of a standard Transformer.

This need for selective, target-specific connectivity has a loose analogue in the brain. White matter forms a dense network of axons between regions, but communication is not uniform: regions have specialized pathways, and ongoing neural activity modulates which connections exert the greatest influence (Appendix~\ref{sec:dti}; \citealp{markov2014,fries2015rhythms}). Motivated by this combination of broad reach and context-dependent selection, we propose \textbf{WhiteMatter} to dynamically channel information across all depths (see Figure~\ref{fig:architecture}). It allows every layer to draw on past-token representations from any depth. For each target layer and past token, a context-dependent router mixes the source depths into a specialized representation, which is then projected into a KV channel. A pool of $k$ channels can be shared across layers, trading finer connection specialization for a smaller KV cache.

These feedback connections are straightforward during decoding, when deeper states of past tokens have already been computed, but their dependencies complicate parallel training and prefill. Feedback Transformer processes tokens
sequentially, limiting GPU parallelism and scalability
~\citep{fan2020feedback,wu-tu-2024-layer}. LCKV uses Jacobi iteration instead: each pass processes all tokens in parallel but can read only the KV produced by the preceding pass~\citep{wu-tu-2024-layer}. Because updates move forward only between passes, convergence requires many repeated evaluations of the full sequence. We introduce cyclic Gauss--Seidel iteration (Figure~\ref{fig:iteration}), which processes strided token groups in order while evaluating the tokens within each group in parallel. Later groups can use updates from earlier groups in the same pass, allowing information to propagate farther with each full-sequence evaluation.

We evaluate both the quality benefit and execution cost of this design.
Under matched training tokens, full-cache
WhiteMatter achieved $7.7\%$ lower perplexity than vanilla with the same number
of layers and performed comparably to vanilla with $50\%$ more layers. Half-cache WhiteMatter
lowers held-out perplexity by $5.9\%$ and $4.3\%$ relative to matched vanilla
models, up to approximately $1.3$B parameters. The
$1.3$B WhiteMatter model has decoding throughput comparable to vanilla's, with $39.4\%$ lower peak device memory and higher
prefill cost.
Controlled studies show that
cyclic iteration converges faster than Jacobi,
and that training closer to convergence improves quality and stability under
further refinement.

\section{Related Work}
\label{sec:related}

\textbf{Deep-to-shallow feedback connections.}
Feedback Transformer pools past-token layer states into a shared
memory~\citep{fan2020feedback}. LCKV uses top-layer KV for condensed layers
and same-depth KV for warm-up layers, with Jacobi iteration for parallel
training~\citep{wu-tu-2024-layer}. T2MLR injects a prior token's middle-layer
state into an earlier layer's residual stream~\citep{cai2026t2mlr}, while
Recurrent Transformer constructs each layer's KV from that layer's past
outputs~\citep{oncescu2026recurrent}. Concurrent with our work, Latent
Recurrent Transformer (LRT)~\citep{huang2026latent} proposes an interleaved
training schedule with a single refinement sweep but does not study how
multiple interleaved passes affect convergence. These methods restrict
feedback to prescribed source layers or a single shared mixture, whereas
WhiteMatter produces content-dependent mixtures specialized for different
layers.

\textbf{Latent reasoning and looping.}
Staircase attention reprocesses past tokens alongside new
tokens~\citep{ju2021staircase}. Coconut and the PonderLM family introduce latent computation through recycled
representations or additional sequence positions~\citep{hao2024training,
  zeng2025pretraining,song2026adaponderlm,zeng2025ponderlm2,li2026ponderlm3}.
These methods add intermittent feedback connections while increasing computation in decoding.

\textbf{Feedforward cross-layer aggregation.}
Several architectures improve information flow through depth by modifying
residual connections, using learned aggregation of earlier-layer
representations~\citep{pagliardini2024denseformer,
  menghani2024laurel,kimiteam2026attention,luo2026delta} or mixing of multiple
residual streams~\citep{zhu2024hyperconnections,xie2025mhc}.
DeepCrossAttention and MUDDFormer form separate mixtures of earlier
representations for attention inputs~\citep{heddes2025deepcrossattention,
  xiao2025muddformer}. Value-residual methods reuse first-layer
values~\citep{zhou2025value,gunasekaran2026satformer}, while KV sharing and
reconstruction methods apply cross-layer reuse to reduce cache
storage~\citep{brandon2024reducing,zuhri2024mlkv,sun2024yoco,
  youwu2024systematic,filippova2026stochastic}.
FusedKV stores KV only in the lower half of the layers and reconstructs each
upper layer's keys and values through separate learned, channel-wise mixtures
of the first and middle layers' caches~\citep{lin2026fusedkv}.
These methods demonstrate the benefit of reusing representations from earlier layers.

\section{Method}
\label{sec:method}

For a Transformer decoder with $L$ layers and width $D$, let $T$ be the sequence
length and $i$, $\ell$, and $j$ index tokens, layers, and channels.
WhiteMatter replaces per-layer KV projections with a \emph{cross-layer KV
  pool}: a router mixes all $L$ states into $k\le L$ channels, each with its
own KV projection. Each layer reads one channel, so the KV cache size is
$k/L$ of vanilla's.

\subsection{Cross-layer KV pool}
\label{sec:method:pool}

\begin{figure}[t]
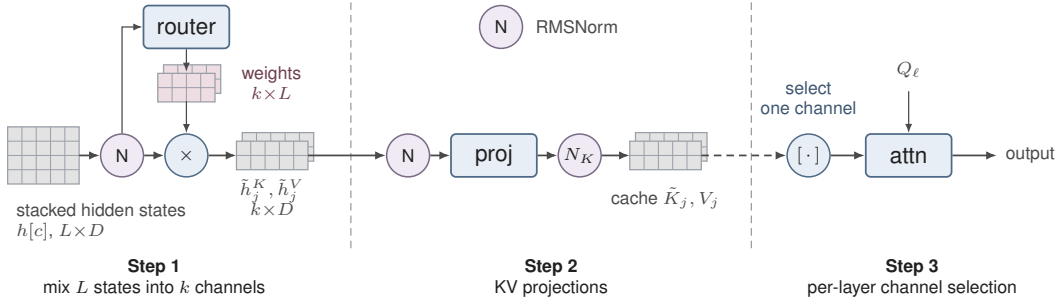

  \vspace{-0.2cm}
  \centering
  \resizebox{\linewidth}{!}{\poolpipelinefigure}
  \vspace{-0.3cm}
  \caption{\textbf{The cross-layer KV pool for one token position $i$.} Dashed
    dividers separate the three steps of \S\ref{sec:method:pool}. In Step 1 a
    data-dependent router mixes the $L$ per-layer states into $k$ shared
    channels. In Step 2 the resulting channels undergo KV projection; K
    normalization and RoPE are then applied to the keys before cache storage.
    In Step 3 each query-side layer reads one channel; the dashed arrow marks
    the cache boundary, as the
    stored channels are read while processing a later token. The key and value
    branches are processed independently.
  }
  \label{fig:pool}
\end{figure}

\textbf{Mixing $L$ states into $k$ channels.} Let $h_\ell[i]\in\mathbb{R}^D$ be the hidden state entering layer $\ell$ at token $i$. At each position $i$, the pool combines the
$L$ source states into $k$ channels using dynamic mixing weights,
computed independently for the key and value branches. We describe the key
branch; the value branch is identical with its own parameters.

Each source state is first RMS-normalized, giving $\hat{h}^K_\ell[i]$. This
pre-mix norm puts the $L$ layers on a common scale and keeps their magnitudes
from growing as they recur through the feedback loop.

The router reads normalized states from every $p$th source layer to reduce
its parameter count, while producing mixing weights over all $L$ layers.
Let $\xi^K[i]$ concatenate the selected states. The mixing weights and resulting channel are
\begin{equation*}
  \alpha^K_j[i] = W^{\alpha K}_j\, \xi^K[i] + b^{\alpha K}_j,
  \qquad
  \tilde{h}^K_j[i] = \sum_{\ell=0}^{L-1} \alpha^K_j[i][\ell]\, \hat{h}^K_\ell[i].
\end{equation*}
The weights are signed and can therefore express differences among layer representations. The value branch uses the same construction with its own norm, router
$W^{\alpha V}_j, b^{\alpha V}_j$, and weights $\alpha^V_j[i]$ applied to
$\hat{h}^V_\ell[i]$.

\textbf{KV projections.} A second RMSNorm places the mixed
channels at a common scale before they are projected into keys and values:
\begin{equation*}
  K_j[i] = W^K_j\, \mathrm{RMSNorm}^K_j(\tilde{h}^K_j[i]),
  \qquad
  V_j[i] = W^V_j\, \mathrm{RMSNorm}^V_j(\tilde{h}^V_j[i]),
\end{equation*}
Per-channel key normalization and RoPE are applied before caching the keys;
values are cached directly. The pool constructs each token's channels
independently from its layer states, so KV can be refreshed at selected
positions without recomputing the remaining cache.

\textbf{Per-layer channel selection.} Layer $\ell$ reads channel
$j=\ell\bmod k$. With $k=L$, each layer receives a distinct mixture;
smaller $k$ shares mixtures across layers and reduces the KV cache to $k/L$
of vanilla's size. Reading all channels would require each layer to stream
all $k$ key and value channels from HBM. The fixed selection preserves one
channel read per layer. Appendix~\ref{sec:pool-details} specifies router
parameterization, initialization, and cache indexing.

\begin{figure}[t]
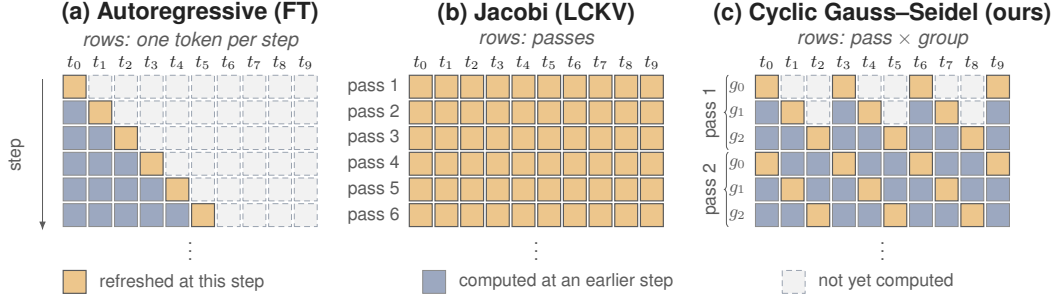

  \vspace{-10px}
  \centering
  \resizebox{\linewidth}{!}{\iterationmodesfigure}
  \vspace{-10px}
  \caption{\textbf{Schedules for resolving the
      feedback connections.} Rows are computation steps and columns are tokens;
    each cell is shaded according to when its KV source was last updated.
    \textbf{(a)} Autoregressive: Feedback Transformer processes each token sequentially.
    \textbf{(b)} Jacobi: LCKV processes all tokens in parallel in multiple passes.
    \textbf{(c)} Cyclic: strided groups run in order, so later
    groups read earlier groups' updates within a pass.}
  \label{fig:iteration}
\end{figure}

\subsection{Autoregressive decoding}
\label{sec:method:decode}

WhiteMatter's KV channels depend on states from all depths. Reading the
current token's channels would therefore create a dependency cycle between
shallow and deep layers. We use \emph{strictly causal attention}: token $i$
attends only to positions $s<i$. After its layer-stack evaluation, we
construct its channels and append them to the cache for subsequent tokens.
Decoding requires one layer-stack evaluation and one pool evaluation per
token; a learned boundary token supplies the initial KV entry.

\subsection{Parallel training and prefill}
\label{sec:method:iteration}

Let $\mathrm{KV}[i]$ denote the key and value channels at position $i$. Let
$\operatorname{Model}(X;\mathrm{KV}_{\mathrm{in}})$ denote one complete model
pass that reads $\mathrm{KV}_{\mathrm{in}}$ and returns the newly constructed
cache $\mathrm{KV}_{\mathrm{out}}$. Because of the strict causality introduced in
\S\ref{sec:method:decode}, the KV produced by exact autoregressive execution
for an input token sequence $X=(x[0],\dots,x[T{-}1])$ satisfies
\begin{equation*}
  \mathrm{KV} = \operatorname{Model}\!\left(X;\mathrm{KV}\right).
\end{equation*}
The cache can therefore be viewed as a fixed point and solved by iteration, thereby reducing the number of sequential operations needed in training and prefill, as shown in Figure~\ref{fig:iteration}.

\paragraph{Jacobi iteration.}
\citet{wu-tu-2024-layer} approximate the fixed point with $n$
token-parallel passes, analogous to the Jacobi method for solving linear systems.
We first initialize $\mathrm{KV}^{(0)}$ from each token's embedding, treating the
embedding as its state at every source layer. For $t=1,\dots,n$, we update
\begin{equation*}
  \mathrm{KV}^{(t)}
  = \operatorname{Model}\!\left(X;\mathrm{KV}^{(t-1)}\right).
\end{equation*}
All $T$ tokens are evaluated in parallel using the previous pass's cache.
This maximizes token parallelism, but updates cannot propagate between
tokens until the next pass, so convergence can require many full-sequence
evaluations.

\textbf{Cyclic Gauss--Seidel iteration.}
To propagate updates within a pass while retaining token parallelism, we
divide token positions into $g$ groups and process groups in sequence;
tokens within each group are evaluated in parallel. We refresh KV after each group so
that later groups can read its updates, analogous to Gauss--Seidel iteration.
Since attention often concentrates on nearby tokens, we use a cyclic group assignment
$\mathcal{G}_q=\{i:i\bmod g=q\}$, evaluated in order $q=0,\ldots,g-1$,
to favor propagation between neighboring positions. For example, each
token in group $q>0$ can read its immediate predecessor's current-pass KV.
Entries from groups not yet processed retain their previous-pass values,
and all reads obey the same strictly causal $s<i$ mask.
Algorithm~\ref{alg:cyclic} gives the complete schedule. We implement a cyclic attention kernel by adapting
FlashAttention-style tiling~\citep{dao2023flashattention2} to the strided
causal mask.

\begin{wrapfigure}{r}{0.6\columnwidth}
  \vspace{-0.7cm}
  \begin{minipage}{\linewidth}
    \begin{algorithm}[H]
      \small
      \caption{Cyclic iteration}
      \label{alg:cyclic}
      \begin{algorithmic}[1]
        \Require Token embeddings $X$, number of passes $n$, number of groups $g$
        \State Initialize $\mathrm{KV}$ from token embeddings
        \For{$t=1,\ldots,n$}
        \For{$q=0,\ldots,g-1$}
        \State $\mathcal{G}_q\gets\{i:0\le i<T,\ i\bmod g=q\}$
        \State Evaluate tokens $\mathcal{G}_q$ in parallel using current $\mathrm{KV}$
        \State Collect their layer inputs $H$ and final outputs $Y[\mathcal{G}_q]$
        \State $\mathrm{KV}[\mathcal{G}_q]\gets\operatorname{Pool}(H)$
        \EndFor
        \EndFor
        \State \Return $Y,\mathrm{KV}$
      \end{algorithmic}
    \end{algorithm}
  \end{minipage}
  \vspace{-0.7cm}
\end{wrapfigure}

Each pass contains $g$ sequential group evaluations, with approximately
$T/g$ tokens evaluated in parallel per group. Increasing $g$ allows updates
to propagate through more groups within a pass, but reduces the parallel
work available in each group. Minimizing execution time therefore requires balancing the number of passes required and the cost of each pass.

\textbf{Truncated backpropagation.} During training, increasing the
number of refinement passes also increases the cost of backpropagation. We follow
\citet{wu-tu-2024-layer} in carrying gradients only through the last
$n_g \le n$ passes; earlier passes run without gradients and
serve to approach the fixed point.

\section{Experiments}
\label{sec:experiments}

\subsection{Experimental setup}
\label{sec:exp:setup}

\textbf{Models and baselines.}
We evaluated Qwen3-based decoders~\citep{qwen3technicalreport} at two scales.
At $D{=}512$, we compared $16$-layer WhiteMatter with full ($k{=}16$) and
half ($k{=}8$) KV caches against vanilla, LCKV~\citep{wu-tu-2024-layer},
and FusedKV~\citep{lin2026fusedkv}. We also trained a $24$-layer vanilla
model to compare against greater depth. At $D{=}1792$, we compared
$28$-layer vanilla and half-cache WhiteMatter ($k{=}14$), with $1.351$B
and $1.326$B parameters, respectively. WhiteMatter used
$g{=}8$ cyclic groups and router stride $p{=}2$.

LCKV used $w\in\{4,7\}$ same-depth warmup layers around a condensed
block sharing one KV source. The bottom/condensed/top layer counts were
$2/12/2$ and $3/9/4$, giving $5/16$ and $1/2$ of vanilla's cache,
respectively; $w{=}7$ matches half-cache WhiteMatter. FusedKV retained KV
in the first eight layers and used layers $0$ and $7$ as sources for the
upper eight layers. Appendix~\ref{sec:experiment-details} gives the remaining
architecture settings.

\textbf{Data and training.}
We trained from scratch on shuffled FineWeb-Edu~\citep{penedo2024fineweb}
with the Qwen3 tokenizer, packing length-$2048$ sequences with EOS separators
and within-document attention. The $D{=}512$ and $D{=}1792$ suites used
matched budgets of $8$B and $10$B tokens and effective batches of $128$ and
$32$, respectively. WhiteMatter used one no-gradient cyclic pass followed
by two gradient-carrying passes. LCKV used seven no-gradient Jacobi passes
followed by two gradient-carrying passes~\citep{wu-tu-2024-layer}; vanilla
and FusedKV used one pass. We used Muon~\citep{jordan2024muon} for
two-dimensional weight matrices and AdamW for the remaining parameters.
Appendix~\ref{sec:experiment-details} gives optimizer and batching details.

\textbf{Quality evaluation.}
We measured perplexity on $5{,}000$ held-out sequences and evaluated
zero-shot benchmarks with lm-evaluation-harness~\citep{eval-harness}.
All evaluations use final checkpoints and full evaluation splits.
WhiteMatter and LCKV use three cyclic and nine Jacobi
passes, respectively, for likelihood evaluation and generation prefill;
generation then uses cached autoregressive decoding. We also evaluate
the perplexity under autoregressive decoding for WhiteMatter and LCKV.
Appendix~\ref{sec:full-downstream} gives task and scoring details.

\textbf{Runtime evaluation.}
We measured throughput and memory for the $1.3$B architectures at batch
$64$ on an RTX A6000 with compiled BF16 execution. Prefill processes
$2048$ prompt tokens; decoding generates $128$ tokens after this prefix.
WhiteMatter uses the evaluation schedule above. LCKV uses nine Jacobi
passes with a $6/15/7$-layer sandwich, and Feedback Transformer processes
prompt tokens autoregressively.
See Appendix~\ref{sec:runtime-details} for measurement details.

\subsection{Language modeling and downstream performance}
\label{sec:exp:main}
\label{sec:evaluation}

\begin{table}[t]
  \vspace{-0.4cm}
  \centering
  \caption{\textbf{Zero-shot evaluation at $D{=}512$ with $8$B training tokens.}
    Both full- and half-cache WhiteMatter models achieved lower perplexity and higher average
    accuracy than all other models, including the 24L vanilla model, which has $50\%$ more layers.
  }
  \label{tab:benchmarks}
  \vspace{4pt}
  \begingroup
  \fontsize{7.5}{9.5}\selectfont
  \setlength{\tabcolsep}{0.35pt}
  \input{data/downstream/benchmark_table.tex}
  \endgroup
  \caption{\textbf{Zero-shot evaluation at $D{=}1792$ with $10$B training tokens.} At $1.3$B total parameters, WhiteMatter also consistently outperforms vanilla while using $50\%$ less KV cache.}
  \label{tab:benchmarks-large}
  \vspace{4pt}
  \begingroup
  \fontsize{7.5}{9.5}\selectfont
  \setlength{\tabcolsep}{0.35pt}
  \input{data/downstream/benchmark_table_large.tex}
  \endgroup
  \begin{minipage}{\linewidth}
    \footnotesize
    In both tables, SQD is SQuAD cloze completion scored by answer containment.
    Avg. is the unweighted mean of the $11$ non-perplexity scores. acc. n denotes length-normalized accuracy and EM denotes exact match.
  \end{minipage}
\end{table}

\begin{figure}[t]
  \vspace{-0.0cm}
  \centering
  \includegraphics[width=0.9\linewidth]{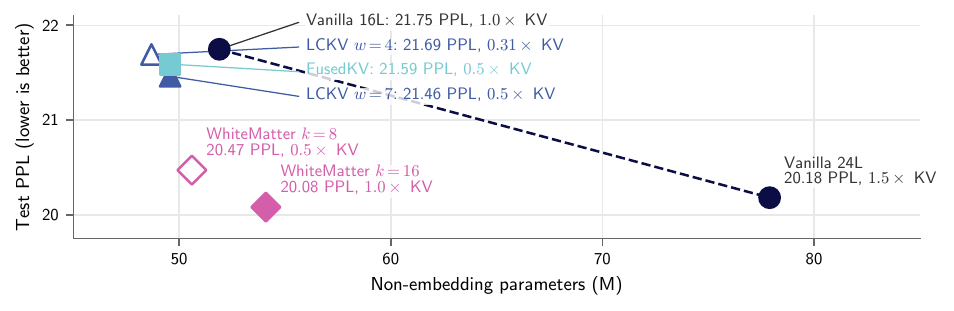}
  \vspace{-0.5cm}
  \caption{\textbf{Language-modeling quality versus model size after $8$B
      training tokens.} Lower perplexity is better; labels indicate KV-cache
    size relative to vanilla 16L. Full-cache WhiteMatter performs comparably to the
    deeper vanilla model, while half-cache WhiteMatter improves over
    same-depth vanilla.}
  \label{fig:main-results}
\end{figure}

\textbf{Language-modeling quality.}
Half-cache WhiteMatter reduces held-out perplexity by $5.9\%$ at the smaller
scale and $4.3\%$ at approximately $1.3$B parameters, relative to matched
vanilla models. At the smaller scale, it also outperforms equal-cache LCKV
and FusedKV, while full-cache WhiteMatter achieves perplexity comparable to
the deeper vanilla model with fewer non-embedding parameters
(Figure~\ref{fig:main-results}). For all evaluated WhiteMatter and LCKV
models, autoregressive perplexity differs by less than $1\%$ from that
obtained with fixed-point iteration.

\textbf{Zero-shot performance.}
Half-cache WhiteMatter improves the average downstream score over matched
vanilla models at both scales (Tables~\ref{tab:benchmarks}
and~\ref{tab:benchmarks-large}). At the smaller scale, both half- and
full-cache WhiteMatter outperform all baselines on average, including
the vanilla model with $50\%$ more layers,
with full-cache WhiteMatter achieving the highest average score.

\subsection{Prefill convergence}
\label{sec:prefill-convergence}

\begin{figure}[t]
  \vspace{-0.0cm}
  \centering
  \begin{minipage}[t]{0.49\linewidth}
    \vspace{0pt}
    \centering
    \includegraphics[width=\linewidth]{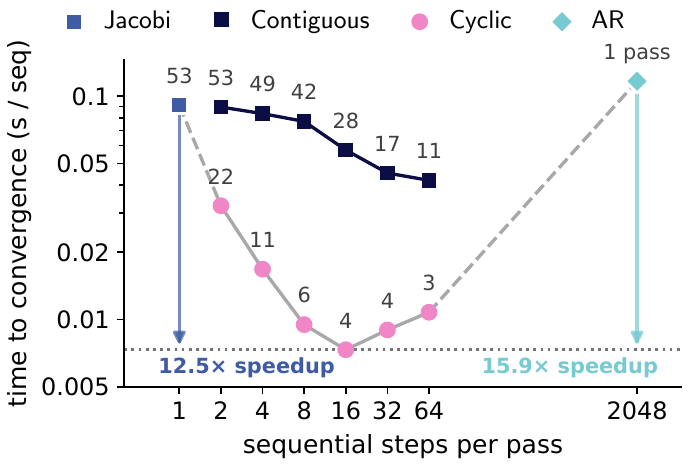}
    \vspace{-22pt}
    \captionof{figure}{\textbf{Prefill runtime at matched PPL.}
      Labels show pass counts; AR denotes autoregressive execution. Increasing
      the number of sequential steps per pass reduces the passes needed but
      increases the cost of each pass. Cyclic iteration balances these
      effects, achieving the lowest runtime and a $12.5\times$ speedup over
      Jacobi.}
    \label{fig:prefill-convergence}
  \end{minipage}\hfill
  \begin{minipage}[t]{0.49\linewidth}
    \vspace{0pt}
    \centering
    \includegraphics[width=\linewidth]{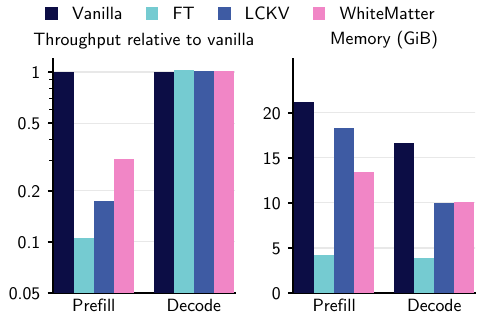}
    \vspace{-22pt}
    \captionof{figure}{\textbf{Inference throughput and peak device memory.}
      Throughput (log scale) is relative to vanilla.
      FT denotes Feedback Transformer.  WhiteMatter achieves decoding throughput comparable to vanilla's with
      lower memory use. Its prefill is slower than vanilla's but faster than
      that of the other feedback architectures.}
    \label{fig:runtime-summary}
  \end{minipage}
\end{figure}

We evaluated prefill convergence on a $4$-layer model
trained with exact autoregressive execution for $78.6$M tokens.
For each schedule, we selected the first pass whose perplexity was at most
$1\%$ above the exact autoregressive reference. Pass counts were selected
in FP32; Figure~\ref{fig:prefill-convergence} reports compiled BF16 decoder
runtime at those counts, using length-$2048$ sequences and batch $32$.
Appendix~\ref{sec:prefill-convergence-details}
gives more detail.

Cyclic $g{=}16$ reaches the target in four passes and takes
$7.32$\,ms/sequence, compared with $53$ passes and $91.20$\,ms/sequence
for Jacobi, a $12.5\times$ speedup. More groups reduce the passes needed
but increase the cost of each pass, so the largest group count does not
give the lowest runtime.

We also evaluated dividing each pass into contiguous chunks processed from left to right. Cyclic groups require fewer passes
and less time at every tested group count, supporting their use for
faster propagation within a pass.

\subsection{Prefill and decoding efficiency}
\label{sec:runtime}

At batch size $64$, WhiteMatter prefill achieves $31\%$ of vanilla's
throughput, $1.78\times$ LCKV's, and $2.92\times$ Feedback Transformer's
(Figure~\ref{fig:runtime-summary}). Its peak device memory is
$13.44$\,GiB, compared with vanilla's $21.21$\,GiB.
All four architectures decode at approximately $2{,}500$ tokens/s.
WhiteMatter uses $10.05$\,GiB of peak device memory versus vanilla's
$16.59$\,GiB, a $39.4\%$ reduction. LCKV uses a similar $10.01$\,GiB,
while Feedback Transformer uses $3.86$\,GiB.

\section{Analysis}

\begin{figure*}[t]
  \vspace{-0.2cm}
  \centering
  \includegraphics[width=\textwidth]{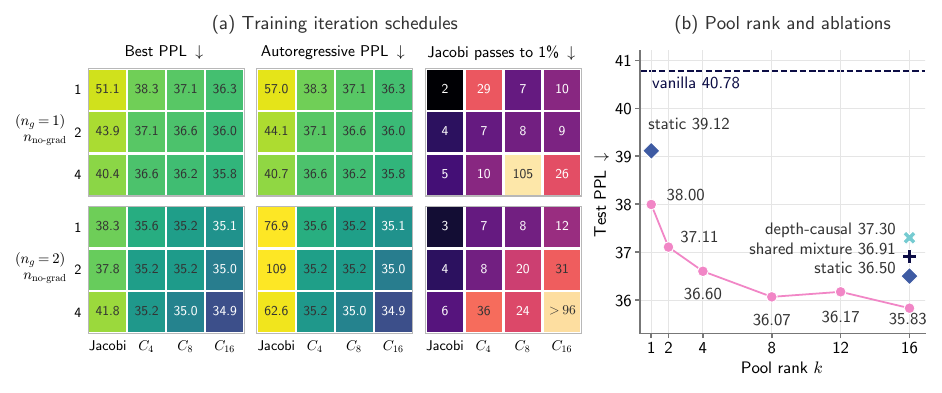}
  \vspace{-0.6cm}
  \caption{
    \textbf{(a) Training schedules.} Training with more refinement generally
    improves perplexity and reduces degradation under further iteration,
    although approaching the best quality can require more inference passes.
    Row groups vary gradient-carrying passes, inner rows vary no-gradient
    passes, and columns compare Jacobi with cyclic schedules
    ($C_g$: $g$ groups).
    \textbf{(b) Channel count and ablations.} More KV channels generally
    improve quality, but increasing cache capacity alone does not recover
    the benefit of specialized source mixtures. Content-dependent routing
    and deep-to-shallow connections also improve performance.}
  \label{fig:analysis-summary}
\end{figure*}

For the schedule and connectivity experiments, we used the $16$-layer,
$D{=}512$ model and the training setup in \S\ref{sec:exp:setup}, reducing
the budget to $20{,}000$ steps at global batch $8$ ($327.7$M tokens).

\textbf{Iteration schedules and convergence.}
We study how training closer to convergence affects model quality.
With $k{=}8$, we varied gradient passes $n_g\in\{1,2\}$, no-gradient passes
$n_{\text{no-grad}}\in\{1,2,4\}$, and schedules
$\{\mathrm{Jacobi},C_4,C_8,C_{16}\}$ over two seeds.
Figure~\ref{fig:analysis-summary}a reports the best perplexity over passes
$1$--$32$ under each training schedule, approximate autoregressive
perplexity after $32 C_{16}$ passes, and the Jacobi passes needed to reach
within $1\%$ of the best perplexity.
The strongest schedule lowers perplexity by $32\%$ relative to the weakest.
More passes and cyclic groups improve quality with diminishing returns.
Models trained farther from the fixed point degrade under further iteration;
those trained closer remain stable after convergence but can require more
Jacobi passes to approach their best perplexity.

\textbf{Computational cost.}
At $D{=}512$, WhiteMatter uses $0.99$--$1.03\times$ vanilla's decoding
FLOPs, but $2.32$--$2.50\times$ its training FLOPs and $3.05$--$3.30\times$
its three-pass prefill FLOPs, excluding the LM head
(Appendix~\ref{sec:flop-details}). Truncated backpropagation limits training
cost, while wall time also depends on parallelism and memory traffic.

We further examine how channel count, distinct source mixtures, content-dependent
routing, and deep-to-shallow access contribute to model quality.

\textbf{Channel count.}
We trained models with $k\in\{1,2,4,8,12,16\}$ (Figure~\ref{fig:analysis-summary}b). More channels generally improve quality,
most strongly from $k{=}1$ to $k{=}2$.

\textbf{Specialized mixtures.}
We replace the distinct source mixtures with one shared mixture while
retaining $k{=}16$ independent KV projections. This model performs worse
than WhiteMatter with $k{=}4$, despite using $4\times$ the cache.

\textbf{Dynamic routing.}
Replacing the dynamic router with static learnable weights raises perplexity
by $3.0\%$ at $k{=}1$ and $1.9\%$ at $k{=}16$.

\textbf{Deep-to-shallow feedback.}
Restricting layer $\ell$ to KV mixtures of states $0,\ldots,\ell$ removes
feedback and iteration. This full-cache model improves over vanilla, but its
perplexity is $4.1\%$ higher than full-cache WhiteMatter, supporting the
benefit of deep-to-shallow reuse.

\section{Conclusion}
We introduced WhiteMatter, which gives Transformer layers access to
past-token representations from all depths through content-dependent KV
source mixing. Distinct mixtures enable specialized connections, while
channel sharing reduces KV-cache storage. Under matched training-token
budgets, half-cache WhiteMatter improved perplexity and average downstream
performance over vanilla at both evaluated scales, up to $1.3$B parameters;
at the smaller scale, full-cache WhiteMatter performs comparably to vanilla with
$50\%$ more layers. Ablations supported distinct mixtures, dynamic routing,
and deep-to-shallow feedback.
We also introduced cyclic Gauss--Seidel iteration to accelerate training
and prefill, reaching the target perplexity $12.5\times$ faster than Jacobi
on the autoregressively trained reference model. At $1.3$B parameters,
WhiteMatter matched vanilla decoding throughput with $39.4\%$ lower peak
device memory, although iterative training and prefill remained more
expensive. Overall, specialized reuse of past-token states improves quality
while reducing cache storage; further reducing training and prefill costs
remains future work.

\section*{Limitations}

Despite speedups relative to autoregressive computation and Jacobi iteration, iterative training and prefill remain more expensive than those of standard Transformers.
Our main quality results cover up to $1.3$B parameters trained for $10$B tokens;
effectiveness on larger models trained with more data requires further investigation. Full-cache and alternative KV-sharing
baselines are evaluated only at the smaller scale.

\subsection*{AI use statement}

We used generative AI tools to implement methods.

We did not use generative AI tools to help develop theoretical models or
conceptual frameworks, formulate mathematical claims, propose or refine
hypotheses, design or provide feedback on research methodology or experiments,
support qualitative and thematic data analysis, or interpret results. The
following uses were not applicable to this work: generating synthetic data
sets; providing critical ingredients for proving mathematical claims;
assisting in the writing of proofs; assisting with translation; and cleaning
and reformatting datasets.

We also used generative AI tools to create or modify scientific figures or
images, create or edit software code, draft parts of the paper, summarize or
analyse existing literature, source or search for information, edit the paper
to improve readability, identify relevant literature, format references, and
suggest a structure for the paper.

We reviewed all AI-assisted work. We performed extensive manual literature
surveys alongside LLM-assisted ones. LLM-generated code was verified and
tested for correctness. The manuscript was manually reviewed and edited in multiple revisions. We take responsibility for the final content
of this work, including text, claims or artifacts produced with the aid of
generative AI.

\input{main.bbl}
\appendix
\raggedbottom
\clearpage

\section{White-matter connectivity}
\label{sec:dti}

\begin{figure}[H]
  \centering
  \includegraphics[height=5.8cm]{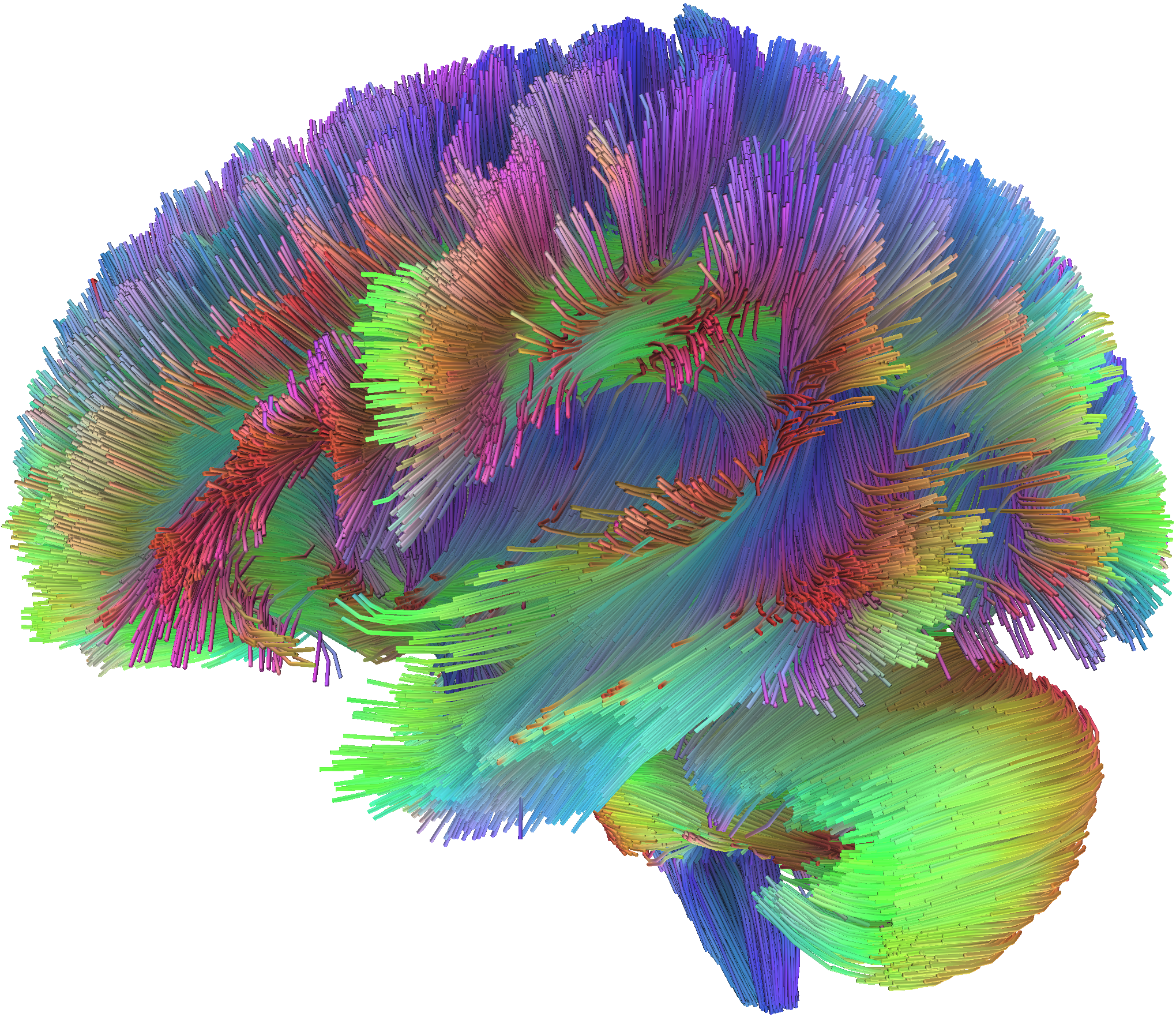}\hfill
  \includegraphics[height=5.8cm]{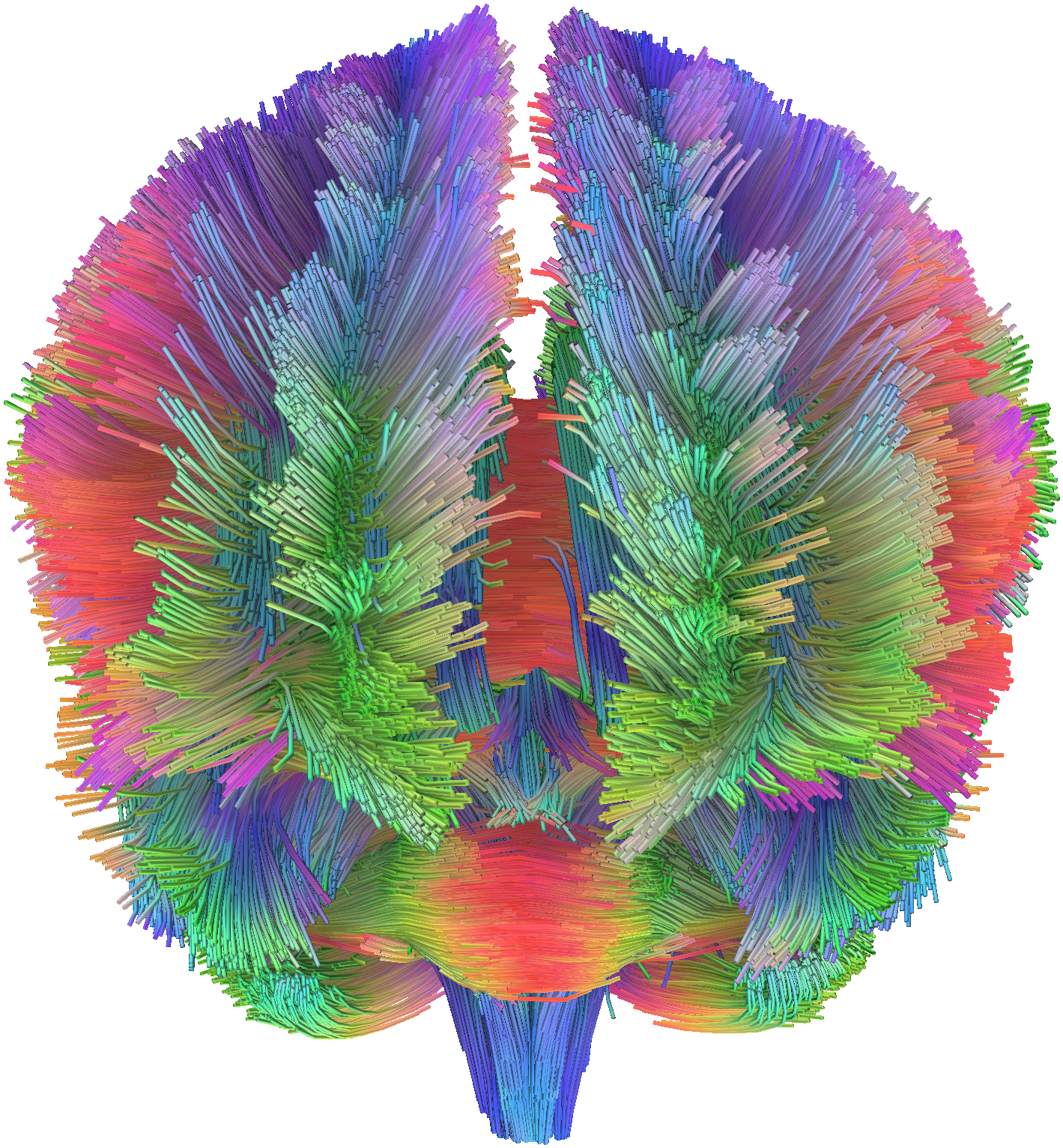}
  \caption{\textbf{Whole-brain white-matter tractography.} A
    population-averaged human structural connectome reconstructed from
    diffusion MRI, rendered as fiber tracts with DSI Studio~\citep{dsistudio}.
    The population-averaged template~\citep{yeh2018template} was built from
    Human Connectome Project data~\citep{vanessen2013hcp}.}
  \label{fig:dti}
\end{figure}

\section{KV pool implementation details}
\label{sec:pool-details}

\paragraph{Router parameterization.}
The selected layers are spaced backward from $L-1$ at stride $p$ and
concatenated in increasing depth order. With $L'=\lceil L/p\rceil$,
$\xi^K[i]\in\mathbb{R}^{L'D}$ and each channel's router has
$W^{\alpha K}_j\in\mathbb{R}^{L\times L'D}$ and
$b^{\alpha K}_j\in\mathbb{R}^{L}$. We stack these parameters into one
linear layer with $kL$ outputs, reshaped to $k\times L$.

\paragraph{Router initialization.}
Router weights start at zero. Biases initially select the top source for
$k=1$, sources satisfying $\ell\bmod k=j$ for $1<k<L$, and source
$\min(j+1,L-1)$ for $k=L$. In the main half-cache and full-cache models,
selected bias entries are $0.25$ and all others are zero.

\paragraph{Projection and cache indexing.}
Each channel has projection matrices
$W^K_j,W^V_j\in\mathbb{R}^{H_{\mathrm{kv}}d\times D}$, where
$H_{\mathrm{kv}}$ is the number of KV heads and $d$ is the head dimension.
The boundary embedding is repeated across source layers before pool
projection; its KV occupies slot zero, and real token $i$ occupies slot $i+1$.
The boundary has rotary position zero. Real-token positions start at one
and reset for each packed document. The document mask always permits the
boundary entry.

\section{Experimental details}
\label{sec:experiment-details}

We used \texttt{karpathy}'s \texttt{fineweb-edu-100b-shuffle} dataset and
the Qwen3-0.6B-Base tokenizer (vocabulary $151{,}936$).
Table~\ref{tab:configurations} gives additional architecture settings.
Full-cache WhiteMatter uses two gradient-accumulation steps with global
microbatch $64$; all other main configurations use one accumulation step.
The $1.3$B parameter counts include the tied embedding/output matrix
($272.3$M parameters) once.

\begin{table}[h]
  \centering
  \caption{\textbf{Additional pretraining settings.}}
  \label{tab:configurations}
  \begin{tabular}{lrr}
    \hline
    Setting            & $D{=}512$  & $D{=}1792$  \\
    \hline
    Intermediate width & $1536$     & $5376$      \\
    Query / KV heads   & $6$ / $3$  & $14$ / $7$  \\
    Head dimension     & $96$       & $128$       \\
    Optimizer steps    & $30{,}518$ & $152{,}588$ \\
    \hline
  \end{tabular}
\end{table}

\paragraph{Optimization.} Muon used momentum $0.95$ and five
Newton--Schulz steps; AdamW used $\beta_1=0.9$ and $\beta_2=0.95$.
Both optimizers used a peak learning rate of
$3{\times}10^{-4}$, $2\%$ warmup, cosine decay to $10\%$ of the peak, and weight
decay of $0.1$. Training used bfloat16
autocast with fp32 master weights on eight NVIDIA RTX
A6000 GPUs. Before DDP all-reduce, each GPU clipped the gradient norm at $1.0$.

\paragraph{Held-out perplexity at the larger scale.}
At approximately $1.3$B parameters, vanilla and half-cache WhiteMatter
achieved held-out perplexities of $13.51$ and $12.93$, respectively, under
the evaluation protocol in \S\ref{sec:exp:setup}.

\section{Inference benchmark details}
\label{sec:runtime-details}

We report median throughput over five runs after compilation and cache
setup. Timings include the output head and greedy token selection;
prefill computes only the final token's logits. Decoding uses CUDA graphs
and excludes prefix preparation.

Peak memory is the maximum of two measurements during steady-state execution:
the sampled device-memory peak, and PyTorch peak reserved memory plus a
stable external-memory baseline.

\section{FLOP measurement details}
\label{sec:flop-details}

We measured FLOPs with PyTorch at batch $1$ and sequence length $2048$,
using causal attention and the schedules in \S\ref{sec:exp:setup}.
Decoding uses $2048$ cached tokens. Counts include two operations per
multiply-add and FusedKV's elementwise cache fusion. FusedKV prefill runs
all prompt tokens through the source layers and only the final token
through the reconstruction layers.

\begin{table}[h]
  \centering
  \caption{\textbf{Per-token FLOPs for training, prefill, and
      decoding at $D{=}512$, in GFLOPs per token.}}
  \label{tab:flops}
  \vspace{4pt}
  \setlength{\tabcolsep}{4.5pt}
  \begin{tabular}{lrrr}
    \hline
    Model                & Decode & Prefill & Training \\
    \hline
    Vanilla 16L          & 0.179  & 0.142   & 0.444    \\
    FusedKV              & 0.203  & 0.071   & 0.429    \\
    LCKV $w{=}4$         & 0.173  & 0.935   & 1.421    \\
    LCKV $w{=}7$         & 0.175  & 0.738   & 1.174    \\
    WhiteMatter $k{=}8$  & 0.177  & 0.432   & 1.028    \\
    WhiteMatter $k{=}16$ & 0.184  & 0.467   & 1.111    \\
    \hline
  \end{tabular}
\end{table}

\section{Convergence measurement details}
\label{sec:convergence-details}

\subsection{Controlled prefill convergence}
\label{sec:prefill-convergence-details}

\paragraph{Reference model and schedules.}
The model in \S\ref{sec:prefill-convergence} has $4$ layers, $D{=}512$, and
$k{=}4$. It was trained with exact autoregressive execution for $800$ steps
at length $1024$ and batch $96$, totaling $78.6$M tokens.
We compared Jacobi iteration with cyclic groups and contiguous
chunks, using $g\in\{2,4,8,16,32,64\}$ for both partitions.
Contiguous chunks are processed from left to right. Both grouped schedules
evaluate tokens within each group in parallel using fixed KV and refresh
the group's KV before processing the next group.

\begin{figure}[ht]
  \centering
  \includegraphics[width=\linewidth]{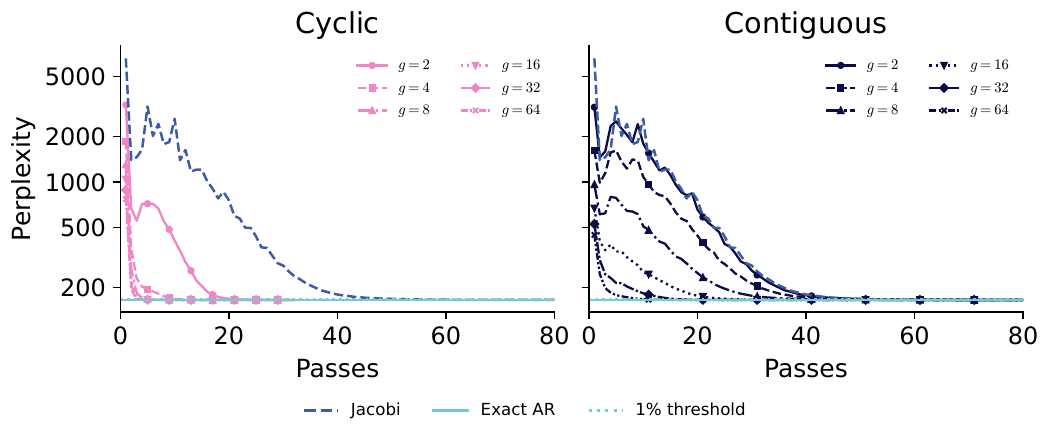}
  \caption{\textbf{FP32 prefill convergence on the autoregressively trained
      reference model.} Both partitions use $g\in\{2,4,8,16,32,64\}$ groups.
    Curves show all evaluated passes. Horizontal lines mark exact
    autoregressive perplexity and the $1\%$ threshold used for pass selection.}
  \label{fig:prefill-trajectories}
\end{figure}

\paragraph{Pass selection.}
We used FP32 weights and arithmetic to evaluate on $192$
unseen single-document sequences of length $2048$. We pooled cross-entropy
over all $393{,}024$ next-token targets before exponentiating; the exact
autoregressive reference perplexity was $165.44$. For each schedule, we
selected the first pass whose perplexity was at most $1.01$ times this
reference. Figure~\ref{fig:prefill-trajectories} shows the convergence trajectories.

\paragraph{Timing.}
Timing uses one A6000 GPU with compiled BF16
execution and batch size $32$. We report median batch time divided
by $32$ after five warmups and $30$ trials, or $10$ full rollouts for
autoregressive evaluation. Timing covers the decoder, excluding embeddings,
final normalization, the output head, and token selection. Contiguous
attention uses FlashAttention~2. Pass counts are selected in FP32; the BF16
timings measure execution cost at those counts.

\section{Convergence of a larger cyclic-trained model}
\label{sec:legacy-prefill-convergence}

\begin{figure}[H]
  \centering
  \includegraphics[width=0.8\linewidth]{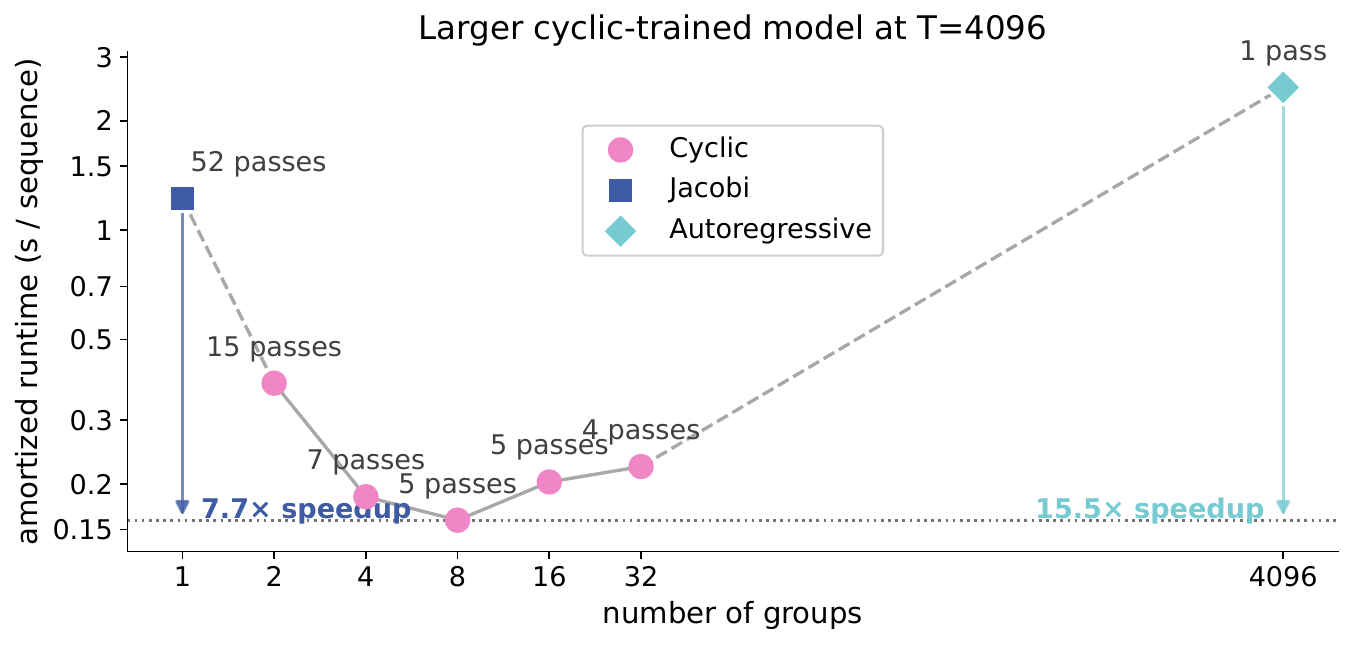}
  \caption{\textbf{Iteration-schedule runtime for a larger cyclic-trained
      model.} Cyclic iteration achieves the lowest runtime at the selected
    pass counts.}
\end{figure}

The $8$-layer, $D{=}1024$, $k{=}8$ model was trained for $122{,}000$
steps at global batch $8$ and length $4096$ (approximately $4.0$B tokens)
with cyclic $g{=}8$ iteration and frozen pretrained embeddings.
Evaluation uses $192$ held-out sequences at length $4096$. We select the
fewest passes reaching within $1\%$ of the fp32 autoregressive perplexity
($18.37$) and measure runtime at batch $64$.

At the selected pass counts, Jacobi takes $1.220$\,s/sequence for $52$
passes and cyclic $g{=}8$ takes $0.159$\,s/sequence for five passes.
The autoregressive rollout takes $2.470$\,s/sequence.

\section{Downstream evaluation details}
\label{sec:full-downstream}

\paragraph{Task scoring.}
In Tables~\ref{tab:benchmarks} and~\ref{tab:benchmarks-large}, acc.\,n denotes
length-normalized accuracy. LAMBADA perplexity is token-based; WikiText
perplexity is word-based. BLiMP averages across $67$ minimal-pair subtasks
($67{,}000$ pairs), comparing grammatical and ungrammatical sentences.
SciQ prepends the supplied support passage and scores four candidate answers.
ReCoRD selects the highest-likelihood candidate completion and normalizes
the predicted and accepted entities before comparison.

\paragraph{SQuAD generation.}
We use \texttt{squad\_completion}~\citep{arora2024simple} with the
$2{,}984$-example validation split of \texttt{hazyresearch/based-squad}.
Answer matching is case-insensitive. Greedy generation permits up to
$256$ new tokens and stops at a blank line or EOS. The adapter retains
up to $768$ prompt tokens within the $1024$-token context limit.
FusedKV recomputes the prefix at each generated step.

\end{document}

%% file: math_commands.tex
\usepackage{amsmath,amsfonts,bm}

\def\eqref#1{equation~\ref{#1}}

\def\1{\bm{1}}

\def\rr{{\textnormal{r}}}
\def\rs{{\textnormal{s}}}

\DeclareMathAlphabet{\mathsfit}{\encodingdefault}{\sfdefault}{m}{sl}
\SetMathAlphabet{\mathsfit}{bold}{\encodingdefault}{\sfdefault}{bx}{n}



%% file: figures/architecture_4panel.tex
\newcommand{\architecturefourpanelfigure}{%
  \begingroup%
  \begin{tikzpicture}[x=1mm,y=1mm]

    \definecolor{layergray}{HTML}{D9D9D9}
    \definecolor{layeredge}{HTML}{8A8A8A}
    \definecolor{alphapink}{HTML}{B36680}
    \definecolor{betabluebase}{HTML}{4D78A8}
    \colorlet{betablue}{betabluebase!80!gray}
    \definecolor{paneltitle}{HTML}{222222}

    \tikzset{
      layer/.style={
          rounded corners=1.1pt, draw=layeredge, line width=0.45pt,
          fill=layergray, minimum width=6mm, minimum height=2.8mm,
          inner sep=0pt
        },
      capsule/.style={
          ellipse, draw=layeredge, line width=0.45pt,
          minimum width=4.8mm, minimum height=3.0mm, inner sep=0pt
        },
      kvcache/.style={
          capsule, fill=alphapink!14, font=\scriptsize\sffamily, text=layeredge
        },
      ffarrow/.style={
          -{Latex[length=0.85mm,width=0.65mm]},
          draw=layeredge!55, line width=0.32pt
        },
      produce/.style={
          -{Latex[length=1.05mm,width=0.8mm]},
          draw=alphapink, line width=0.48pt
        },
      consume/.style={
          -{Latex[length=1.05mm,width=0.8mm]},
          draw=betablue, line width=0.45pt
        },
      paneltitle/.style={font=\scriptsize\bfseries\sffamily, text=paneltitle},
      subtitle/.style={font=\scriptsize\itshape\sffamily, text=paneltitle},
      tinylabel/.style={font=\scriptsize\sffamily, text=paneltitle},
    }

    \def\nL{6}
    \def\rowsep{4.2}
    \def\kvx{10}
    \def\dstx{20}
    \def\panelstep{35}

    \begin{scope}[shift={(0,0)}]
      \foreach \ell in {0,...,5} {
          \pgfmathsetmacro{\yc}{(\nL-\ell)*\rowsep}
          \node[layer] (V-src-\ell) at (0,\yc) {};
          \node[kvcache] (V-kv-\ell) at (\kvx,\yc) {};
          \node[layer] (V-dst-\ell) at (\dstx,\yc) {};
          \draw[produce] (V-src-\ell.east) -- (V-kv-\ell.west);
          \draw[consume] (V-kv-\ell.east) -- (V-dst-\ell.west);
        }
      \foreach \side in {src,dst} {
          \foreach \ell in {1,...,5} {
              \pgfmathtruncatemacro{\upper}{\ell-1}
              \draw[ffarrow] (V-\side-\ell.north) -- (V-\side-\upper.south);
            }
        }
      \node[paneltitle] at (\kvx,35.2) {(a) Vanilla};
      \node[subtitle] at (\kvx,31.4) {same-depth KV};
      \node[tinylabel] at (0,0.5) {past token};
      \node[tinylabel] at (\dstx,0.5) {current token};
    \end{scope}

    \begin{scope}[shift={(\panelstep,0)}]
      \foreach \ell in {0,...,5} {
          \pgfmathsetmacro{\yc}{(\nL-\ell)*\rowsep}
          \node[layer] (FB-src-\ell) at (0,\yc) {};
          \node[layer] (FB-dst-\ell) at (\dstx,\yc) {};
        }
      \node[kvcache] (FB-mem) at (\kvx,14.7) {};
      \foreach \ell in {0,...,5} {
          \pgfmathsetmacro{\op}{0.38+0.10*\ell}
          \draw[produce,opacity=\op] (FB-src-\ell.east) -- (FB-mem.west);
          \draw[consume,opacity=0.72] (FB-mem.east) -- (FB-dst-\ell.west);
        }
      \foreach \side in {src,dst} {
          \foreach \ell in {1,...,5} {
              \pgfmathtruncatemacro{\upper}{\ell-1}
              \draw[ffarrow] (FB-\side-\ell.north) -- (FB-\side-\upper.south);
            }
        }
      \node[paneltitle] at (\kvx,35.2) {(b) Feedback Transformer};
      \node[subtitle] at (\kvx,31.4) {one static shared mixture};
      \node[tinylabel] at (0,0.5) {past token};
      \node[tinylabel] at (\dstx,0.5) {current token};
    \end{scope}

    \begin{scope}[shift={(2*\panelstep,0)}]
      \foreach \ell in {0,...,5} {
          \pgfmathsetmacro{\yc}{(\nL-\ell)*\rowsep}
          \node[layer] (LC-src-\ell) at (0,\yc) {};
          \node[layer] (LC-dst-\ell) at (\dstx,\yc) {};
        }

      \foreach \ell in {0,5} {
          \pgfmathsetmacro{\yc}{(\nL-\ell)*\rowsep}
          \node[kvcache] (LC-warm-kv-\ell) at (\kvx,\yc) {};
          \draw[produce] (LC-src-\ell.east) -- (LC-warm-kv-\ell.west);
          \draw[consume]
          (LC-warm-kv-\ell.east) -- (LC-dst-\ell.west);
        }

      \node[kvcache] (LC-condensed-kv) at (\kvx,14.7) {};
      \draw[produce] (LC-src-1.east) -- (LC-condensed-kv.west);
      \foreach \ell in {1,...,4} {
          \draw[consume,opacity=0.72]
          (LC-condensed-kv.east) -- (LC-dst-\ell.west);
        }

      \foreach \side in {src,dst} {
          \foreach \ell in {1,...,5} {
              \pgfmathtruncatemacro{\upper}{\ell-1}
              \draw[ffarrow] (LC-\side-\ell.north) -- (LC-\side-\upper.south);
            }
        }
      \node[paneltitle] at (\kvx,35.2) {(c) LCKV};
      \node[subtitle] at (\kvx,31.4) {top condensed-layer KV};
      \node[tinylabel] at (0,0.5) {past token};
      \node[tinylabel] at (\dstx,0.5) {current token};
    \end{scope}

    \begin{scope}[shift={(3*\panelstep,0)}]
      \def\nK{3}
      \foreach \ell in {0,...,5} {
          \pgfmathsetmacro{\yc}{(\nL-\ell)*\rowsep}
          \node[layer] (WM-src-\ell) at (0,\yc) {};
          \node[layer] (WM-dst-\ell) at (\dstx,\yc) {};
        }
      \foreach \j in {0,1,2} {
          \pgfmathsetmacro{\yc}{14.7+(1-\j)*6.2}
          \node[kvcache] (WM-ch-\j) at (\kvx,\yc) {};
        }

      \foreach \j in {0,1,2} {
          \foreach \ell in {0,...,5} {
              \pgfmathsetmacro{\wmw}{0.20+0.07*mod(2*\ell+3*\j,6)}
              \pgfmathsetmacro{\wmo}{0.38+0.09*mod(\ell+2*\j,6)}
              \draw[-{Latex[length=1.0mm,width=0.75mm]},draw=alphapink,
                line width=\wmw pt,opacity=\wmo]
              (WM-src-\ell.east) -- (WM-ch-\j.west);
            }
        }
      \foreach \ell in {0,...,5} {
          \pgfmathtruncatemacro{\j}{mod(\ell,\nK)}
          \draw[consume] (WM-ch-\j.east) -- (WM-dst-\ell.west);
        }
      \foreach \side in {src,dst} {
          \foreach \ell in {1,...,5} {
              \pgfmathtruncatemacro{\upper}{\ell-1}
              \draw[ffarrow] (WM-\side-\ell.north) -- (WM-\side-\upper.south);
            }
        }
      \node[paneltitle] at (\kvx,35.2) {(d) \textbf{WhiteMatter} (ours)};
      \node[subtitle] at (\kvx,31.4) {dynamic all-depth channels};
      \node[tinylabel] at (0,0.5) {past token};
      \node[tinylabel] at (\dstx,0.5) {current token};
    \end{scope}

    \foreach \i in {1,2,3} {
        \pgfmathsetmacro{\separatorx}{
          \i*\panelstep-(\panelstep-\dstx)/2}
        \draw[gray!42,line width=0.3pt]
        (\separatorx,1.8) -- (\separatorx,36.2);
      }

    \node[layer] at (0,-3.3) {};
    \node[tinylabel,anchor=west,inner sep=0pt] at (5,-3.3) {block};
    \node[kvcache] at (24,-3.3) {};
    \node[tinylabel,anchor=west,inner sep=0pt] at (28,-3.3) {stored KV};
    \draw[ffarrow] (47,-3.3) -- (52,-3.3);
    \node[tinylabel,anchor=west,inner sep=0pt] at (54,-3.3) {hidden state};
    \draw[produce] (73,-3.3) -- (78,-3.3);
    \node[tinylabel,anchor=west,inner sep=0pt] at (80,-3.3) {source to KV};
    \draw[consume] (103,-3.3) -- (108,-3.3);
    \node[tinylabel,anchor=west,inner sep=0pt] at (110,-3.3) {KV consumption};
  \end{tikzpicture}
  \endgroup
}

%% file: figures/pool_pipeline.tex
\providecolor{layergray}{HTML}{E3E3E3}
\providecolor{layeredge}{HTML}{9AA0A8}
\providecolor{alphapink}{HTML}{B36680}
\providecolor{betablue}{HTML}{4D78A8}
\providecolor{modedge}{HTML}{4F6A86}
\providecolor{modfill}{HTML}{E8EEF5}
\providecolor{normfill}{HTML}{EFEAF3}
\providecolor{normedge}{HTML}{7A6A86}
\providecolor{arcdark}{HTML}{4A4A4A}
\providecolor{paneltitle}{HTML}{1F1F1F}

\tikzset{
  mod/.style   = {rounded corners=2pt, draw=modedge, line width=0.7pt,
      fill=modfill, align=center, font=\small\sffamily,
      text=paneltitle, inner sep=1.5pt, minimum width=12mm,
      minimum height=6mm},
  norm/.style  = {circle, draw=normedge, line width=0.6pt, fill=normfill,
      inner sep=0pt, minimum size=6mm, font=\scriptsize\sffamily,
      text=paneltitle},
  matmul/.style= {circle, draw=modedge, line width=0.6pt, fill=modfill,
      inner sep=0pt, minimum size=6mm, font=\scriptsize\sffamily,
      text=paneltitle},
  selop/.style = {matmul},
  side/.style   = {font=\scriptsize\sffamily\bfseries, text=paneltitle, align=center},
  shp/.style   = {font=\scriptsize\sffamily, text=arcdark, align=center},
  stepdiv/.style = {densely dashed, draw=arcdark!55, line width=0.5pt},
  steplab/.style = {font=\scriptsize\sffamily, text=paneltitle, align=center,
      inner sep=0pt},
  flow/.style  = {-{Latex[length=1.5mm,width=1.2mm]}, draw=arcdark, line width=0.7pt},
  flowcache/.style = {flow, densely dashed},
  ctrl/.style  = {-{Latex[length=1.3mm,width=1.0mm]}, draw=arcdark, line width=0.5pt},
}

\def\cellsz{2.0}
\newcommand{\tgrid}[5]{%
  \pgfmathsetmacro{\gw}{#3*\cellsz}\pgfmathsetmacro{\gh}{#4*\cellsz}%
  \pgfmathtruncatemacro{\cm}{#3-1}\pgfmathtruncatemacro{\rm}{#4-1}%
  \fill[#5] (#1-\gw/2,#2-\gh/2) rectangle (#1+\gw/2,#2+\gh/2);%
  \ifnum\cm>0\foreach \cc in {1,...,\cm}{\draw[layeredge,line width=0.15pt] (#1-\gw/2+\cc*\cellsz,#2-\gh/2)--(#1-\gw/2+\cc*\cellsz,#2+\gh/2);}\fi%
  \ifnum\rm>0\foreach \rr in {1,...,\rm}{\draw[layeredge,line width=0.15pt] (#1-\gw/2,#2-\gh/2+\rr*\cellsz)--(#1+\gw/2,#2-\gh/2+\rr*\cellsz);}\fi%
  \draw[layeredge,line width=0.5pt] (#1-\gw/2,#2-\gh/2) rectangle (#1+\gw/2,#2+\gh/2);%
}
\newcommand{\tgridstack}[5]{%
  \tgrid{#1+0.8}{#2+1.2}{#3}{#4}{#5}%
  \tgrid{#1}{#2}{#3}{#4}{#5}%
}

\newcommand{\poolpipelinefigure}{%
  \begin{tikzpicture}[x=1mm, y=1mm]

    \def\xH{0}\def\xNa{11}\def\xAmix{20}\def\xHt{32}\def\xNb{51}%
    \def\xProj{63}\def\xNk{75}\def\xCache{87}\def\xBread{107}%
    \def\xAttn{121}\def\xOut{138}

    \draw[stepdiv] (43,-13) -- (43,22);
    \draw[stepdiv] (99,-13) -- (99,22);
    \node[steplab, anchor=north] at (15.5,-14.5)
    {\textbf{Step 1}\\mix $L$ states into $k$ channels};
    \node[steplab, anchor=north] at (71,-14.5)
    {\textbf{Step 2}\\KV projections};
    \node[steplab, anchor=north] at (121.5,-14.5)
    {\textbf{Step 3}\\per-layer channel selection};

    \tgrid{\xH}{0}{5}{4}{layergray}
    \node[shp, anchor=north west, align=left] at (\xH-5,-5.5) {stacked hidden states\\$h[c]$, $L{\times}D$};

    \node[norm] (n1) at (\xNa,0) {N};

    \node[matmul] (amix) at (\xAmix,0) {$\times$};
    \node[mod] (router) at (\xAmix,18) {router};
    \tgridstack{\xAmix}{9.5}{4}{2}{alphapink!22}
    \node[shp, anchor=west, text=alphapink!60!black] at (\xAmix+6.5, 10) {weights\\$k{\times}L$};

    \tgridstack{\xHt}{0}{5}{2}{layergray}
    \node[shp] at (\xHt,-6) {$\tilde h^K_j,\tilde h^V_j$\\$k{\times}D$};

    \node[norm] (n2) at (\xNb,0) {N};

    \node[mod] (proj) at (\xProj,0) {proj};

    \node[norm] (nk) at (\xNk,0) {$N_K$};

    \tgridstack{\xCache}{0}{5}{2}{layergray}
    \node[shp] at (\xCache,-6) {cache $\tilde K_j, V_j$};

    \node[selop] (bread) at (\xBread,0) {$[\,\cdot\,]$};
    \node[shp, anchor=south, text=betablue!70!black] at (\xBread, 4.5) {select\\one channel};

    \node[mod] (attn) at (\xAttn,0) {attn};
    \node[shp] at (\xAttn,12) {$Q_\ell$};

    \node[shp] (out) at (\xOut,0) {output};

    \node[norm] (lgN) at (64,18) {N};
    \node[shp, anchor=west] at (67.6,18) {RMSNorm};

    \draw[flow] (\xH+5,0)  -- (n1.west);
    \draw[flow] (n1.east)  -- (amix.west);
    \draw[flow] (amix.east)-- (\xHt-5,0);
    \draw[flow] (\xHt+5,0) -- (n2.west);
    \draw[flow] (n2.east)  -- (proj.west);
    \draw[flow] (proj.east)-- (nk.west);
    \draw[flow] (nk.east)  -- (\xCache-5,0);
    \draw[flowcache] (\xCache+5,0) -- (bread.west);
    \draw[flow] (bread.east)  -- (attn.west);
    \draw[flow] (attn.east)   -- (out.west);

    \draw[ctrl] (n1.north) |- (router.west);
    \draw[ctrl] (router.south) -- (\xAmix,11.7);
    \draw[ctrl] (\xAmix,7.6) -- (amix.north);
    \draw[ctrl] (\xAttn,9) -- (attn.north);

  \end{tikzpicture}%
}

%% file: figures/iteration_modes.tex
\definecolor{imEdge}{HTML}{8A8A8A}
\definecolor{imFinal}{HTML}{9CA8C0}     
\definecolor{imCurr}{HTML}{EDC68C}      
\definecolor{imEmpty}{HTML}{F2F2F2}     
\definecolor{imStaleEdge}{HTML}{9AA3B0}
\definecolor{imPanel}{HTML}{222222}
\definecolor{imArc}{HTML}{5A5A5A}

\newcommand{\iterationmodesfigure}{%
  \begin{tikzpicture}[x=1mm, y=1mm]
    \tikzset{
      cell/.style    = {draw=imEdge, line width=0.4pt, minimum size=4.1mm,
          inner sep=0pt, anchor=center},
      sNow/.style    = {cell, fill=imCurr, draw=imArc, line width=0.6pt}, 
      sDone/.style   = {cell, fill=imFinal},                              
      sEmpty/.style  = {cell, fill=imEmpty, draw=imStaleEdge, densely dashed},
      toklbl/.style  = {font=\small\sffamily, text=imPanel},
      rowlbl/.style  = {font=\small\sffamily, text=imArc},
      passlbl/.style = {font=\small\sffamily, text=imArc},
      paneltitle/.style = {font=\large\bfseries\sffamily, text=imPanel},
      subt/.style    = {font=\normalsize\itshape\sffamily, text=imArc},
      pcap/.style    = {font=\small\sffamily, text=imArc, align=center},
      brace/.style   = {decorate, decoration={brace, amplitude=2pt, mirror},
          draw=imArc, line width=0.5pt},
    }

    \def\cs{4.6}     
    \def\rs{4.6}     
    \def\panelw{62}  
    \def\midx{4.5*\cs}

    \begin{scope}[shift={(0,0)}]
      \foreach \c in {0,...,9}{ \node[toklbl] at (\c*\cs, 0.95*\rs) {$t_{\c}$}; }
      \foreach \r in {0,...,5}{
          \foreach \c in {0,...,9}{
              \ifnum\c<\r   \node[sDone]  at (\c*\cs, -\r*\rs) {};
              \else\ifnum\c=\r \node[sNow] at (\c*\cs, -\r*\rs) {};
                \else         \node[sEmpty] at (\c*\cs, -\r*\rs) {};
                \fi\fi
            }
        }
      \node[rowlbl, anchor=north, inner sep=1pt] at (\midx, -5.5*\rs) {$\vdots$};
      \draw[-{Latex[length=1.6mm,width=1.2mm]}, draw=imArc, line width=0.5pt]
      (-1.25*\cs, 0.3*\rs) -- (-1.25*\cs, -5.6*\rs);
      \node[rowlbl, rotate=90, anchor=south] at (-1.7*\cs, -2.65*\rs) {step};
      \node[paneltitle, anchor=south] at (\midx, 2.15*\rs) {(a) Autoregressive (FT)};
      \node[subt, anchor=center] at (\midx, 1.8*\rs) {rows: one token per step};
    \end{scope}

    \begin{scope}[shift={(\panelw,0)}]
      \foreach \c in {0,...,9}{ \node[toklbl] at (\c*\cs, 0.95*\rs) {$t_{\c}$}; }
      \foreach \r in {0,...,5}{
          \foreach \c in {0,...,9}{ \node[sNow] at (\c*\cs, -\r*\rs) {}; }
          \pgfmathtruncatemacro{\pp}{\r+1}
          \node[passlbl, anchor=east] at (-0.55*\cs, -\r*\rs) {pass \pp};
        }
      \node[rowlbl, anchor=north, inner sep=1pt] at (\midx, -5.5*\rs) {$\vdots$};
      \node[paneltitle, anchor=south] at (\midx, 2.15*\rs) {(b) Jacobi (LCKV)};
      \node[subt, anchor=center] at (\midx, 1.8*\rs) {rows: passes};
    \end{scope}

    \begin{scope}[shift={(2*\panelw,0)}]
      \foreach \c in {0,...,9}{ \node[toklbl] at (\c*\cs, 0.95*\rs) {$t_{\c}$}; }
      \foreach \S in {0,...,5}{
          \foreach \c in {0,...,9}{
              \pgfmathtruncatemacro{\q}{mod(\c,3)}
              \pgfmathtruncatemacro{\ci}{mod(\S,3)}
              \ifnum\q=\ci      \node[sNow]  at (\c*\cs, -\S*\rs) {};
              \else\ifnum\q<\S  \node[sDone] at (\c*\cs, -\S*\rs) {};
                \else             \node[sEmpty]at (\c*\cs, -\S*\rs) {};
                \fi\fi
            }
          \pgfmathtruncatemacro{\gi}{mod(\S,3)}
          \node[rowlbl, anchor=east] at (-0.55*\cs, -\S*\rs) {$g_{\gi}$};
        }
      \draw[brace] (-1.5*\cs, 0.45*\rs) -- (-1.5*\cs, -2.45*\rs);
      \node[passlbl, rotate=90, anchor=center] at (-2.17*\cs, -1.0*\rs) {pass 1};
      \draw[brace] (-1.5*\cs, -2.55*\rs) -- (-1.5*\cs, -5.45*\rs);
      \node[passlbl, rotate=90, anchor=center] at (-2.17*\cs, -4.0*\rs) {pass 2};
      \node[rowlbl, anchor=north, inner sep=1pt] at (\midx, -5.5*\rs) {$\vdots$};
      \node[paneltitle, anchor=south] at (\midx, 2.15*\rs) {(c) Cyclic Gauss--Seidel (ours)};
      \node[subt, anchor=center] at (\midx, 1.8*\rs) {rows: pass $\times$ group};
    \end{scope}

    \begin{scope}[shift={(0, -7.6*\rs)}]
      \node[sNow]   (lf) at (0,0) {};
      \node[rowlbl, anchor=west] at (0.7*\cs, 0) {refreshed at this step};
      \node[sDone]  (ls) at (14*\cs,0) {};
      \node[rowlbl, anchor=west] at (14*\cs+0.7*\cs, 0) {computed at an earlier step};
      \node[sEmpty] (le) at (28*\cs,0) {};
      \node[rowlbl, anchor=west] at (28*\cs+0.7*\cs, 0) {not yet computed};
    \end{scope}

  \end{tikzpicture}%
}

%% file: data/downstream/benchmark_table.tex
\begin{tabular*}{\linewidth}{@{\extracolsep{\fill}}lW{c}{0.32in}W{c}{0.32in}@{\hspace{2.5pt}\vrule\hspace{2.5pt}}W{c}{0.32in}W{c}{0.32in}W{c}{0.32in}W{c}{0.32in}W{c}{0.32in}W{c}{0.32in}W{c}{0.32in}W{c}{0.32in}W{c}{0.32in}W{c}{0.32in}W{c}{0.32in}@{\hspace{0.25pt}\vrule\hspace{0.25pt}}W{c}{0.32in}@{}}
\hline\noalign{\vskip 3pt}
\textbf{Model} & {\fontsize{7}{8}\selectfont\bfseries Wiki.} & {\fontsize{7}{8}\selectfont\bfseries LMB.} & {\fontsize{7}{8}\selectfont\bfseries LMB.} & {\fontsize{7}{8}\selectfont\bfseries BLiMP} & {\fontsize{7}{8}\selectfont\bfseries PIQA} & {\fontsize{7}{8}\selectfont\bfseries Hella.} & {\fontsize{7}{8}\selectfont\bfseries ARC-E} & {\fontsize{7}{8}\selectfont\bfseries ARC-C} & {\fontsize{7}{8}\selectfont\bfseries Wino.} & {\fontsize{7}{8}\selectfont\bfseries OBQA} & {\fontsize{7}{8}\selectfont\bfseries SciQ} & {\fontsize{7}{8}\selectfont\bfseries ReC.} & {\fontsize{7}{8}\selectfont\bfseries SQD} & {\fontsize{7}{8}\selectfont\bfseries Avg.} \\
 & PPL $\downarrow$ & PPL $\downarrow$ & acc. $\uparrow$ & acc. $\uparrow$ & acc. $\uparrow$ & acc.\,n $\uparrow$ & acc. $\uparrow$ & acc.\,n $\uparrow$ & acc. $\uparrow$ & acc.\,n $\uparrow$ & acc. $\uparrow$ & EM $\uparrow$ & acc. $\uparrow$ & acc. $\uparrow$ \rule[-1.5pt]{0pt}{1.5pt} \\
\noalign{\vskip 3pt}\hline\noalign{\vskip 3pt}
Vanilla 16L & 49.34 & 127.47 & 24.43 & 77.23 & 62.73 & 31.67 & 52.15 & 25.00 & 50.04 & 29.00 & 78.50 & 58.50 & 30.03 & 47.21 \\
FusedKV & 48.35 & 92.33 & 27.13 & 76.64 & \textbf{63.71} & 32.25 & 50.76 & 25.00 & \underline{52.01} & \underline{30.20} & 80.30 & 58.96 & 34.82 & 48.34 \\
LCKV $w{=}4$ & 48.81 & 107.52 & 26.33 & 76.17 & 62.73 & 32.52 & 51.43 & 23.72 & 50.91 & \textbf{31.20} & 76.60 & 58.04 & \textbf{45.48} & 48.65 \\
LCKV $w{=}7$ & 49.02 & 102.97 & 25.87 & 75.73 & 61.97 & 32.40 & 52.44 & \textbf{25.94} & \textbf{52.41} & 30.00 & 76.50 & 58.34 & \underline{44.00} & 48.69 \\
WhiteMatter $k{=}8$ & \underline{44.40} & \underline{71.58} & \underline{29.65} & \underline{78.83} & 63.22 & \underline{33.61} & \underline{52.65} & 25.26 & 51.38 & 29.60 & \underline{80.50} & \underline{60.24} & 36.70 & \underline{49.24} \\
WhiteMatter $k{=}16$ & \textbf{43.28} & \textbf{60.73} & \textbf{30.60} & \textbf{79.88} & \underline{63.33} & \textbf{33.80} & \textbf{52.69} & \underline{25.43} & 51.38 & 29.40 & \textbf{83.00} & \textbf{61.52} & 37.80 & \textbf{49.89} \rule[-1.5pt]{0pt}{1.5pt} \\
\noalign{\vskip 3pt}\hline\noalign{\vskip 3pt}
Vanilla 24L & 44.71 & 97.40 & 26.90 & 77.73 & 63.49 & 33.21 & 53.16 & 26.37 & 50.99 & 31.80 & 78.20 & 59.40 & 32.54 & 48.52 \rule[-1.5pt]{0pt}{1.5pt} \\
\noalign{\vskip 3pt}\hline
\end{tabular*}
\par\vspace{6pt}

%% file: data/downstream/benchmark_table_large.tex
\begin{tabular*}{\linewidth}{@{\extracolsep{\fill}}lW{c}{0.32in}W{c}{0.32in}@{\hspace{2.5pt}\vrule\hspace{2.5pt}}W{c}{0.32in}W{c}{0.32in}W{c}{0.32in}W{c}{0.32in}W{c}{0.32in}W{c}{0.32in}W{c}{0.32in}W{c}{0.32in}W{c}{0.32in}W{c}{0.32in}W{c}{0.32in}@{\hspace{0.25pt}\vrule\hspace{0.25pt}}W{c}{0.32in}@{}}
\hline\noalign{\vskip 3pt}
\textbf{Model} & {\fontsize{7}{8}\selectfont\bfseries Wiki.} & {\fontsize{7}{8}\selectfont\bfseries LMB.} & {\fontsize{7}{8}\selectfont\bfseries LMB.} & {\fontsize{7}{8}\selectfont\bfseries BLiMP} & {\fontsize{7}{8}\selectfont\bfseries PIQA} & {\fontsize{7}{8}\selectfont\bfseries Hella.} & {\fontsize{7}{8}\selectfont\bfseries ARC-E} & {\fontsize{7}{8}\selectfont\bfseries ARC-C} & {\fontsize{7}{8}\selectfont\bfseries Wino.} & {\fontsize{7}{8}\selectfont\bfseries OBQA} & {\fontsize{7}{8}\selectfont\bfseries SciQ} & {\fontsize{7}{8}\selectfont\bfseries ReC.} & {\fontsize{7}{8}\selectfont\bfseries SQD} & {\fontsize{7}{8}\selectfont\bfseries Avg.} \\
 & PPL $\downarrow$ & PPL $\downarrow$ & acc. $\uparrow$ & acc. $\uparrow$ & acc. $\uparrow$ & acc.\,n $\uparrow$ & acc. $\uparrow$ & acc.\,n $\uparrow$ & acc. $\uparrow$ & acc.\,n $\uparrow$ & acc. $\uparrow$ & EM $\uparrow$ & acc. $\uparrow$ & acc. $\uparrow$ \rule[-1.5pt]{0pt}{1.5pt} \\
\noalign{\vskip 3pt}\hline\noalign{\vskip 3pt}
Vanilla 28L & 26.61 & 23.84 & 39.65 & 80.83 & \textbf{70.46} & 49.80 & 65.74 & 32.59 & 55.49 & 37.60 & 85.90 & 70.25 & 44.00 & 57.48 \\
WhiteMatter $k{=}14$ & \textbf{24.67} & \textbf{18.42} & \textbf{43.80} & \textbf{81.38} & 69.53 & \textbf{52.72} & \textbf{67.21} & \textbf{36.01} & \textbf{57.38} & \textbf{37.80} & \textbf{88.70} & \textbf{72.75} & \textbf{44.34} & \textbf{59.24} \rule[-1.5pt]{0pt}{1.5pt} \\
\noalign{\vskip 3pt}\hline
\end{tabular*}
\par\vspace{6pt}